%% file: naacl-main.tex
\documentclass[11pt,a4paper]{article}
\usepackage[hyperref]{naaclhlt2019}
\usepackage{times}
\usepackage{latexsym}
\usepackage{graphicx}
\usepackage{multirow}
\usepackage{subfig}
\usepackage{float}
\usepackage{amssymb}
\usepackage{algorithm}
\usepackage[noend]{algpseudocode}

\usepackage{xcolor}
\usepackage{pgfplots}
\usepackage{tikz}
\usetikzlibrary{shapes,arrows}
\usetikzlibrary{positioning}
\usepackage{tkz-graph}
\usetikzlibrary{patterns}

\usepackage{url}

\aclfinalcopy 

\title{Multi-task Learning for Multi-modal\\ Emotion Recognition and Sentiment Analysis}

\author{Md Shad Akhtar$^\dagger$, Dushyant Singh Chauhan$^\dagger$, Deepanway Ghosal$^\dagger$, Soujanya Poria$^+$, \\
\textbf{Asif Ekbal$^\dagger$ and Pushpak Bhattacharyya$^\dagger$} \\
  $^\dagger$ Department of Computer Science \& Engineering \\
  Indian Institute of Technology Patna, India \\
  {\tt \{shad.pcs15, 1821CS17, deepanway.ee14, asif, pb\}@iitp.ac.in} \\
  $^+$ School of Computer Science and Engineering, Nanyang Technological University, Singapore \\
  {\tt sporia@ntu.edu.sg }}

\date{}

\begin{document}
\maketitle
\begin{abstract}
  Related tasks often have inter-dependence on each other and perform better when solved in a joint framework. In this paper, we present a deep multi-task learning framework that jointly performs sentiment and emotion analysis both. The multi-modal inputs (i.e., \textit{text}, \textit{acoustic} and \textit{visual frames}) of a video convey diverse and distinctive information, and usually do not have equal contribution in the decision making. We propose a context-level inter-modal attention framework for simultaneously predicting the sentiment and expressed emotions of an utterance. We evaluate our proposed approach on CMU-MOSEI dataset for multi-modal sentiment and emotion analysis. Evaluation results suggest that multi-task learning framework offers improvement over the single-task framework. The proposed approach reports new state-of-the-art performance for both sentiment analysis and emotion analysis.
\end{abstract}

\input{body}

\bibliography{Reference}
\bibliographystyle{acl_natbib}

\end{document}

%% file: body.tex
\section{Introduction}\label{sec:intro}
With the rapid growth of social media video platforms such as Youtube, Vimeo, users now tend to upload videos on these platforms. Such video platforms offer users an opportunity to express their opinions on any topic. Videos usually consist of \textit{audio} and \textit{visual} modalities, and thus can be considered as a source of multi-modal information. Although videos contain more information than text, fusing multiple modalities is a major challenge. A common practice in sentiment analysis and emotion recognition or affective computing, in general, is to analyze textual opinions. However, in recent days multi-modal affect analysis has gained a major attention \cite{poria2017context,poria2016convolutional}.
In these works, in addition to the \textit{visual frames}, other sources of information such as \textit{acoustic} and \textit{textual} (transcript) representation of the spoken languages are also incorporated in the analysis. 
Multi-modal analysis (e.g. sentiment analysis \citealt{zadeh2018acl}, emotion recognition \citealt{poria2016convolutional}, question-answering \citealt{qa:multimodal} etc.) is an emerging field of study, that utilizes multiple information sources for solving a problem. These sources (e.g., text, visual, acoustic, etc.) offer a diverse and often distinct piece of information that a system can leverage on. For example, \textit{`text'} carries semantic information of the spoken sentence, whereas `\textit{acoustic}' information reveals the emphasis (pitch, voice quality) on each word. In contrast, the `\textit{visual}' information (image or video frame) extracts the gesture and posture of the speaker.

Traditionally, `\textit{text}' has been the key factor in any Natural Language Processing (NLP) tasks including sentiment and emotion analysis. However, with the recent emergence of social media platforms and their available multi-modal contents, an interdisciplinary study involving \textit{text}, \textit{acoustic} and \textit{visual} features have drawn significant interest among the research community. Effectively fusing this diverse information is non-trivial and poses several challenges to the underlying problem.

In our current work, we propose a multi-task model to extract both sentiment (i.e. \textbf{\em positive} or \textbf{\em negative}) and emotion (i.e. \textbf{\em anger}, \textbf{\em disgust}, \textbf{\em fear}, \textbf{\em happy}, \textbf{\em sad} or \textbf{\em surprise}) of a speaker in a video. In multi-task framework, we aim to leverage the inter-dependence of these two tasks to increase the confidence of individual task in prediction. For e.g., information about \textit{anger} emotion can help in prediction of \textit{negative} sentiment and vice-versa.

A speaker can utter multiple utterances (a unit of speech bounded by breathes or pauses) in a single video and these utterances can have different sentiments and emotions. We hypothesize that the sentiment (or, emotion) of an utterance often has inter-dependence on other contextual utterances i.e. the knowledge of sentiment (or, emotion) for an utterance can assist in classifying its neighbor utterances. We utilize all three modalities (i.e. \textit{text}, \textit{acoustic} and \textit{visual}) for the analysis. Although all these sources of information are crucial, they are not equally beneficial for each individual instance. Few examples are presented in Table \ref{tab:exm}. In the first example, \textit{visual frames} provide important clues than \textit{textual} information for finding the sentiment of a sarcastic sentence ``\textit{Thanks for putting me on hold! I've all the time in the world.}". Similarly, the \textit{textual} representation of second example ``\textit{I'm fine.}' does not reveal the exact emotion of a \textit{sad} person. For this particular case, \textit{acoustic} or \textit{visual} information such as low tone voice, facial expression etc. have bigger role to play for the classification.  
\begin{table}[h!]
\centering
\resizebox{0.47\textwidth}{!}{
\begin{tabular}{p{13em}|c|lll}
\textbf{Utterance} & Feeling & \textbf{T} & \textbf{A} & \textbf{V} \\ \hline \hline
\textit{Thanks for putting me on hold! I've all the time in the world.} & Sentiment (Negative) & - & - & \checkmark \\ \hline
\textit{I'm fine.} & Emotion (Sad) & - & \checkmark & \checkmark \\
\end{tabular}
}
\caption{Contributing modalities for different scenario. Tick represents the most contributing information.}
\label{tab:exm}
\end{table}

\section{Problem Definition}
Multi-task learning paradigm provides an efficient platform for achieving generalization. Multiple tasks can exploit the inter-relatedness for improving individual performance through a shared representation. Overall, it provides three basic advantages over the single-task learning paradigm a). it helps in achieving generalization for multiple tasks; b). each task improves its performance in association with the other participating tasks; and c). offers reduced complexity because a single system can handle multiple problems/tasks at the same time.   

Sentiments \cite{Panget.al2005} and emotions \cite{ekman1999} are closely related. Most of the emotional states have clear distinction of being a positive or negative situation. Emotional states e.g. `\textit{anger}', `\textit{fear}', `\textit{disgust}', `\textit{sad}' etc. belong to negative situations, whereas `\textit{happy}' and `\textit{surprise}' reflect the positive situations. Motivated by the association of sentiment \& emotion and the advantages of the multi-task learning paradigm, we present a multi-task framework that jointly learns and classifies the sentiments and emotions in a video. 
As stated earlier, contextual-utterances and/or multi-modal information provide important cues for the classification. Our proposed approach applies attention over both of these sources of information simultaneously (i.e., contextual utterance and inter-modal information), and aims to reveal the most contributing features for the classification. We hypothesize that applying attention to contributing neighboring utterances and/or multi-modal representations may assist the network to learn in a better way. 

Our proposed architecture employs a recurrent neural network based contextual inter-modal attention framework. In our case, unlike the previous approaches, that simply apply attention over the contextual utterance for classification, we take a different approach. Specifically, we attend over the contextual utterances by computing correlations among the modalities of the target utterance and the context utterances. This particularly helps us to distinguish which modalities of the relevant contextual utterances are more important for the classification of the target utterance. The model facilitates this modality selection process by attending over the contextual utterances and thus generates better multi-modal feature representation when these modalities from the context are combined with the modalities of the target utterance.
We evaluate our proposed approach on the recent benchmark dataset of CMU-MOSEI \cite{zadeh2018acl}. It is the largest available dataset (approx. 23K utterances) for multi-modal sentiment and emotion analysis (c.f. Dataset Section). The evaluation shows that contextual inter-modal attention framework attains better performance than the state-of-the-art systems for various combinations of input modalities.

The main contributions of our proposed work are three-fold: \textbf{a)} \textit{we leverage the inter-dependence of two related tasks (i.e. sentiment and emotion) in improving each other’s performance using an effective multi-modal framework}; \textbf{b)} \textit{we propose contextual inter-modal attention mechanism that facilitates the model to assign weightage to the contributing contextual utterances and/or to different modalities simultaneously}. Suppose, to classify an utterance `\textit{u1}' of 5 utterances video, visual features of `\textit{u2}' \& `\textit{u4}', acoustic features of `\textit{u3}' and textual features of `\textit{u1}', `\textit{u3}' \& `\textit{u5}' are more important than others. Our attention model is capable of highlighting such diverse contributing features; and \textbf{c)} \textit{we present the state-of-the-arts for both sentiment and emotion predictions}.

\section{Related Work}
\label{sec:lit}
A survey of the literature suggests that multi-modal sentiment prediction is a relatively new area as compared to textual based sentiment prediction \cite{DBLP:conf/icmi/MorencyMD11,poria2017context,zadeh2018multi-sdk}. A good review covering the literature from uni-modal analysis to multi-modal analysis is presented in \cite{poria2017review}. 

\newcite{zadeh:mosi} introduced the multi-modal dictionary to understand the interaction between facial gestures and spoken words better when expressing sentiment. In another work, \newcite{zadeh2017tensor} proposed a Tensor Fusion Network (TFN) model to learn the intra-modality and inter-modality dynamics of the three modalities (i.e., text, visual and acoustic). Authors reported improved accuracy using multi-modality on the CMU-MOSI dataset. These works did not take contextual information into account. \newcite{poria2017context} proposed a Long Short Term Memory (LSTM) based framework for sentiment classification that leverages the contextual information to capture the inter-dependencies between the utterances. \newcite{zadeh2018multi-sdk} proposed multi-attention blocks (MAB) to capture the information across the three modalities (text, visual and acoustic) for predicting the sentiments. Authors evaluated their approach on the different datasets and reported improved accuracies in the range of 2-3\% over the state-of-the-art models. \newcite{W18-3301} proposed a multi-modal fusion model that exclusively uses high-level visual and acoustic features for sentiment classification.

An application of multi-kernel learning based fusion technique was proposed in \cite{poria2016convolutional}, where the authors employed deep convolutional neural network (CNN) for extracting the textual features and fused it with other modalities (\textit{visual} \& \textit{acoustic}) for emotion prediction. \newcite{ranganathan2016multimodal} proposed a convolutional deep belief network (CDBN) models for multi-modal emotion recognition. The author used CDBN to learn salient multi-modal (acoustic and visual) features of low-intensity expressions of emotions. \newcite{hazarika2018self} introduced a self- attention mechanism for multi-modal emotion detection by feature level fusion of text and speech. Recently, \newcite{zadeh2018acl} introduced the CMU-MOSEI dataset for multi-modal sentiment analysis and emotion recognition. They effectively fused the tri-modal inputs through a dynamic fusion graph and also reported competitive performance \textit{w.r.t.} various state-of-the-arts on MOSEI dataset for both sentiment and emotion classification. 

The main difference between the proposed and existing methods is contextual inter-modal attention. Systems \cite{poria2016convolutional,zadeh:mosi,zadeh2017tensor,W18-3301} do not consider context for the prediction. System \cite{poria2017context} uses contextual information for the prediction but without any attention mechanism. In contrast, \cite{zadeh2018multi-sdk} uses multi-attention blocks but did not account for contextual information. Our proposed model is novel in the sense that our approach applies attention over multi-modal information of the contextual utterances in a single step. Thus, it ensures to reveal the contributing features across \textit{multiple modalities} and \textit{contextual utterances} simultaneously for sentiment and emotion analysis. Further, to the best of our knowledge, this is the first attempt at solving the problems of multi-modal sentiment and emotion analysis together in a multi-task framework.  

The contextual inter-modal attention mechanism is not much explored in NLP domains as such. We found one work that accounts for bi-modal attention for visual question-answering (VQA) \cite{qa:multimodal}. However, its attention mechanism differs from our proposed approach in the following manner: a) VQA proposed question guided image-attention, but 
our attention mechanism attends multi-modalities; b) attention is applied over different positions of the image, whereas our proposed approach applies attention over multiple utterances and two-modalities at a time; c). our proposed attention mechanism attends a sequence of utterances (text, acoustic or visual), whereas VQA applies attention in the spatial domain.
In another work, \newcite{ghosal-EtAl:2018:EMNLP} proposed an inter-modal attention framework for the multi-modal sentiment analysis. However, the key differences with our current work are as follows: a) \newcite{ghosal-EtAl:2018:EMNLP} addressed only sentiment analysis, whereas, in our current work, we address both the sentiment and emotion analysis; b) \newcite{ghosal-EtAl:2018:EMNLP} handles only sentiment analysis in single task learning framework, whereas our proposed approach is based on multi-task learning framework, where we solve two tasks, i.e., sentiment analysis and emotion analysis, together in a single network; c) we perform detailed comparative analysis over the single-task vs. multi-task learning; and d) we present state-of-the-art for both sentiment and emotion analysis.     

\section{Multi-task Multi-modal Emotion Recognition and Sentiment Analysis}
\label{sec:method}
\begin{figure*}[!ht]
\begin{center}
\includegraphics[width=0.99\textwidth]{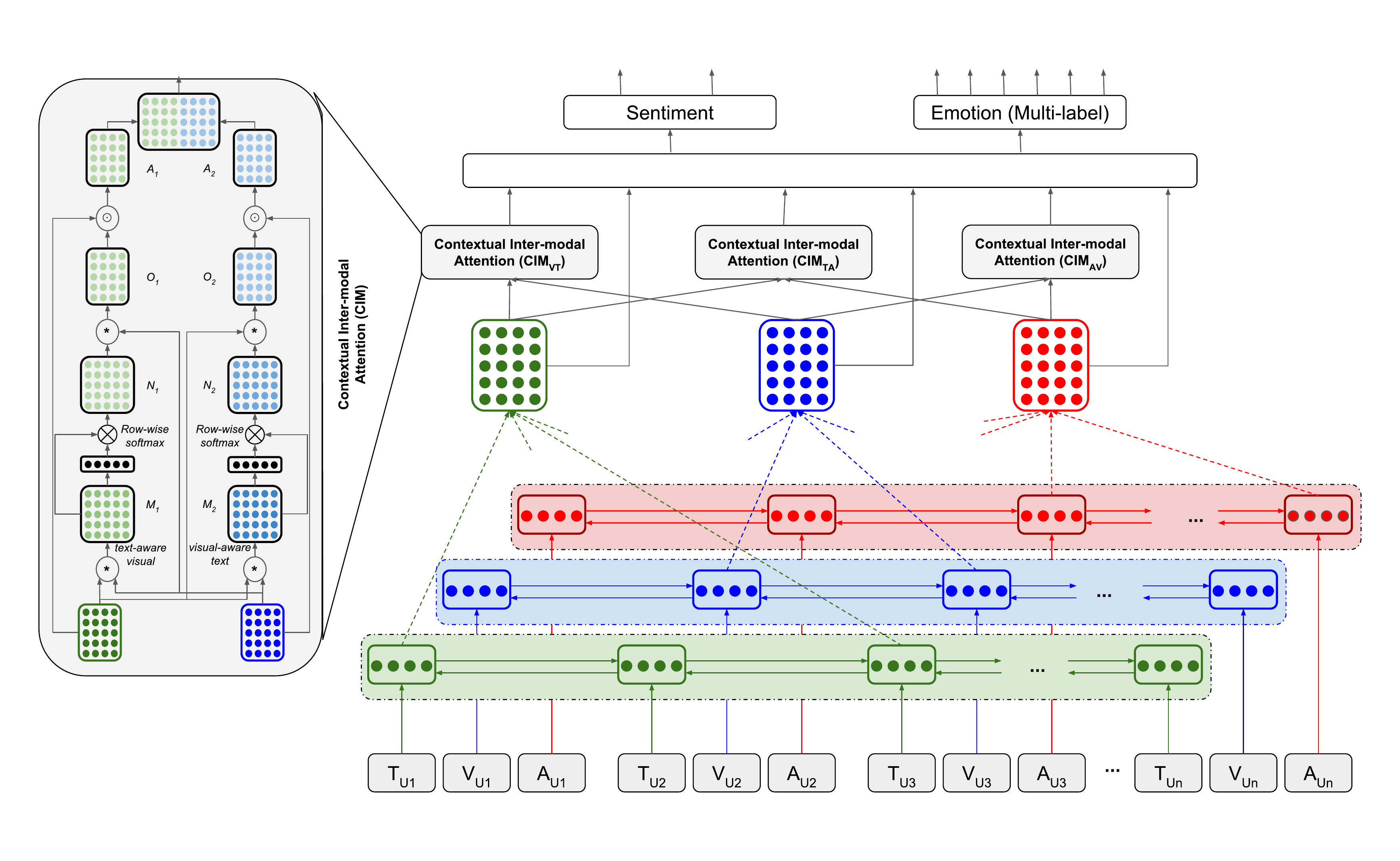}     
\caption{Overall architecture of the proposed framework. Contextual inter-modal (CIM) attention computation between \textit{visual} and \textit{text} modality.}
\label{fig:arch}
\end{center} 
\end{figure*}

In our proposed framework, we aim to leverage multi-modal and contextual information for predicting sentiment and emotion of an utterance simultaneously in a multi-task learning framework. 
As stated earlier, a video consists of a sequence of utterances and their semantics often have inter-dependencies on each other. We employ three bi-directional Gated Recurrent Unit (bi-GRU) network for capturing the contextual information (i.e., one for each modality). Subsequently, we introduce pair-wise inter-modal attention mechanism (i.e. \textit{visual-text}, \textit{text-acoustic} and \textit{acoustic-visual}) to learn the joint-association between the multiple modalities \& utterances. The objective is to emphasize on the contributing features by putting more attention to the respective utterance and neighboring utterances. Motivated by the residual skip connection \cite{he2016deep} the outputs of pair-wise attentions along with the representations of individual modalities are concatenated. Finally, the concatenated representation is shared across the two branches of our proposed network- corresponding to two tasks, i.e., sentiment and emotion classification for prediction (one for each task in the multi-task framework). Sentiment classification branch contains a \textit{softmax} layer for final classification (i.e. \textit{positive} \& \textit{negative}), whereas for emotion classification we use sigmoid layer. The shared representation will receive gradients of error from both the branches (sentiment \& emotion) and accordingly adjust the weights of the models. Thus, the shared representations will not be biased to any particular task, and it will assist the model in achieving generalization for the multiple tasks. Empirical evidences support our hypothesis (c.f. Table \ref{tab:mtl:stl}).

\subsection{Contextual Inter-modal (CIM) Attention Framework}
\label{mmmuba}
Our contextual inter-modal attention framework works on a pair of modalities. At first, we capture the cross-modality information by computing a pair of matching matrices $M_1, M_2 \in \mathbb{R}^{u \times u}$, where \textit{`u'} is the number of utterances in the video. Further, to capture the contextual dependencies, we compute the probability distribution scores ($N_1, N_2 \in \mathbb{R}^{u \times u}$) over each utterance of cross-modality matrices $M_1, M_2$ using a softmax function. This essentially computes the attention weights for contextual utterances. Subsequently, we apply soft attention over the contextual inter-modal matrices to compute the modalitiy-wise attentive representations ($O_1 \& O_2$). Finally, a multiplicative gating mechanism \cite{dhingra2016gated} ($A_1 \& A_2$) is introduced to attend the important components of multiple modalities and utterances. The concatenated attention matrix of $A_1 \& A_2$ then acts as the output of our contextual inter-modal attention framework. The entire process is repeated for each pair-wise modalities i.e. \textit{text-visual}, \textit{acoustic-visual} and \textit{text-acoustic}. We illustrate and summarize the proposed methodology in Figure \ref{fig:arch} and Algorithm \ref{alg:method}, respectively. 

\begin{algorithm}[ht]
\caption{Multi-task Multi-modal Emotion and Sentiment (MTMM-ES)}\label{alg:method}
\begin{algorithmic}
\Procedure{MTMM-ES}{$t,v,a$}
\State $d \gets 100$ \Comment{GRU dimension}
\State $T \gets biGRU_T(t, d)$
\State $V \gets biGRU_V(v, d)$
\State $A \gets biGRU_A(a, d)$
\State $Atn_{TV} \gets$ \textit{CIM-Attention}$(T,V)$
\State $Atn_{AV} \gets$ \textit{CIM-Attention}$(A,V)$
\State $Atn_{TA} \gets$ \textit{CIM-Attention}$(T,A)$
\State $Rep \gets [Atn_{TV}, Atn_{AV}, Atn_{TA}, T, V, A]$
\State $polarity \gets Sentiment(Rep)$
\State $emotion \gets Emotion(Rep)$
\State \textbf{return} $polarity, emotion$
\EndProcedure
\\

\Procedure{CIM-Attention}{$X,Y$}
\State /*Cross-modality information*/
\State $M_1 \gets X.Y^T$
\State $M_2 \gets Y.X^T$
\State /*Contextual Inter-modal attention*/
\For{$i,j \in 1,...,u$} \Comment{$u= \#utterances$}
\State $N_1(i,j) \gets \frac{e^{M_1(i,j)}}{\sum_{k=1}^{u} e^{M_1(i,k)}}$
\State $N_2(i,j) \gets \frac{e^{M_2(i,j)}} {\sum_{k=1}^{u} e^{M_2(i,k)}}$
\EndFor
\State $O_1 \gets N_1.Y$
\State $O_2 \gets N_2.X$
\State /*Multiplicative gating*/
\State $A_1 \gets O_1\odot X$ \Comment{Element-wise mult.}
\State $A_2 \gets O_2\odot Y$

\State \textbf{return} [$A_1, A_2$]
\EndProcedure

\end{algorithmic}
\end{algorithm}

\section{Datasets, Experiments, and Analysis}
\label{sec:exp}
In this section, we describe the datasets used for our experiments and report the results along with necessary analysis. 
\subsection{Datasets}
\label{sec:dataset}
We evaluate our proposed approach on the benchmark datasets of sentiment and emotion analysis, namely CMU Multi-modal Opinion Sentiment and Emotion Intensity (CMU-MOSEI) dataset \cite{zadeh2018acl}. CMU-MOSEI dataset consists of 3,229 videos spanning over 23,000 utterances from more than 1,000 online YouTube speakers. The training, validation \& test set comprises of 16216, 1835 \& 4625 utterances, respectively. 

\input{dataset}
\begin{table}[!ht]
\centering
\resizebox{0.4\textwidth}{!}{
\begin{tabular}{|l|c|l|c|}
\hline
Single emotion & 11050 & Two emotions & 5526\\ \hline
Three emotions & 2084 & Four emotions & 553 \\ \hline
Five emotions & 84 & Six emotions & 8 \\ \hline
No emotion & 3372 \\ \cline{1-2}
\end{tabular}}
\caption{Statistics of multi-label emotions.}
\label{tab:multi-label}
\end{table}

Each utterance has six emotion values associated with it, representing the degree of emotion for \textit{anger}, \textit{disgust}, \textit{fear}, \textit{happy}, \textit{sad} and \textit{surprise}. Emotion labels for an utterance are identified as all non-zero intensity values, i.e. if an utterance has three emotions with non-zero values, we take all three emotions as multi-labels. Further, an utterance that has no emotion label represents the absence of emotion. For experiments, we adopt 7-classes (6 \textit{emotions} + 1 \textit{no emotion}) and pose it as multi-label classification problem, where we try to optimize the binary-cross entropy for each of the class. A brief statistics for multi-label emotions is presented in Table \ref{tab:multi-label}.
In contrast, the sentiment values for each utterance are disjoint, i.e. \textit{value} $< 0$ and \textit{value} $\geq 0$ represent the negative and positive sentiments, respectively. A detailed statistics of the CMU-MOSEI dataset is shown in Table~\ref{tab:dataset}.

\input{results-multi-single}

\subsection{Feature extraction}
We use the CMU-Multi-modal Data SDK\footnote{\url{https://github.com/A2Zadeh/CMU-MultimodalDataSDK}} for downloading and feature extraction. The dataset was pre-tokenized and a feature vector was provided for each word in an utterance. The \textit{textual}, \textit{visual} and \textit{acoustic} features were extracted by \textit{GloVe} \cite{2014glove}, \textit{Facets}\footnote{\url{https://pair-code.github.io/facets/}} \& \textit{CovaRep} \cite{covarep:speech}, respectively. 
Thereafter, we compute the average of \textit{word-level} features to obtain the \textit{utterance-level} features. 

\subsection{Experiments}
We evaluate our proposed approach on the datasets of CMU-MOSEI. We use the Python based Keras library for the implementation. We compute \textit{F1-score} and \textit{accuracy values} for sentiment classification and \textit{F1-score} and \textit{weighted accuracy} \cite{P17-1142} for emotion classification. Weighted accuracy as a metric is chosen due to unbalanced samples across various emotions and it is also in line with the other existing works \cite{zadeh2018acl}. 
To obtain multi-labels for emotion classification, we use 7 sigmoid neurons (corresponds to 6 emotions + 1 no-emotion) with binary cross-entropy loss function. Finally, we take all the emotions whose respective values are above a $threshold$. We optimize and cross-validate both the evaluation metrics (i.e. F1- score and weighted accuracy) and set the \textit{threshold} as  
$0.4$ \& $0.2$ for F1-score and weighted accuracy, respectively. We show our model configurations in Table \ref{tab:param}. 

\begin{table}[ht!]
\centering
\resizebox{0.5\textwidth}{!}
{\begin{tabular}{l|c}
\textbf{Parameters} & \textbf{Values} \\ \hline \hline
Bi-GRU & 2$\times$200 \textit{neurons}, \textit{dropout=0.3} \\
Dense layer & 100 \textit{neurons}, \textit{dropout=0.3} \\
Activations &  \textit{ReLu} \\
Optimizer & \textit{Adam} (\textit{lr=0.001}) \\
Output & \textit{Softmax} (Sent) \& \textit{Sigmoid} (Emo) \\
\multirow{2}{*}{Loss} & \textit{Categorical cross-entropy} (Sent)\\
&  \textit{Binary cross-entropy} (Emo) \\
Threshold & 0.4 (F1) \& 0.2 (W-Acc) for multi-label\\ 
Batch & \textit{16} \\
Epochs & \textit{50} \\
\end{tabular}}
\caption{Model configurations}
\label{tab:param}
\end{table} 

\input{results}
\input{table-attention.tex}
\input{all_attention}
\input{results-compare-new}
As stated earlier, our proposed approach requires at least two modalities to compute bi-modal attention. Hence, we experiment with bi-modal and tri-modal input combinations for the proposed approach i.e. taking \textit{text-visual}, \textit{text-acoustic}, \textit{acoustic-visual} and \textit{text-visual-acoustic} at a time. 
For completeness (i.e., uni-modal inputs), we also experiment with a variant of the proposed approach where we apply self-attention on the utterances of each modality separately. The self-attention unit utilizes the contextual information of the utterances (i.e., it receives \textit{u$\times$d} hidden representations), applies attention and forward it to the output layer for classification. We report the experimental results of both single-task (STL) and multi-task (MTL) learning framework in Table \ref{tab:mtl:stl}. In the single-task framework, we build separate systems for sentiment and emotion analysis, whereas in multi-task framework a joint-model is learned for both of these problems. For sentiment classification, our single-task framework reports an F1-score of 77.67\% and accuracy value of 79.8\% for the tri-modal inputs. Similarly, we obtain 77.71\% F1-score and 60.88\% weighted accuracy for emotion classification.

Comparatively, when both the problems are learned and evaluated in a multi-task learning framework, we observe performance enhancement for both sentiments as well as emotion classification. MTL reports 78.86\% F1-score and 80.47\% accuracy value in comparison to 77.67\% and 79.8\% of STL with tri-modal inputs, respectively. For emotion classification, we also observe an improved F-score (78.6 (MTL) vs. 77.7 (STL)) and weighted accuracy (62.8 (MTL) vs. 60.8 (STL)) in the multi-task framework. It is evident from Figure \ref{mtl-stl-graph} that multi-task learning framework successfully leverages the inter-dependence of both the tasks in improving the overall performance in comparison to single-task learning. The improvements of MTL over STL framework is also statistically significant with \textit{p}-value $<0.05$ (c.f. Table \ref{tab:results-compare}).

We also present attention heatmaps of the multi-task learning framework in Figure \ref{fig:heatmap}. For illustration, we take the video of the first utterance of Table \ref{tab-heatmap-video-example}. It has total six utterances. We depict three pair-wise attention matrices of $2\times(6\times6)$ dimension-one each for \textit{text-visual}, \textit{text-acoustics} and \textit{acoustics-visual}. Solid lines in between represent the boundary of the two modalities, e.g. left side of  Figure \ref{fig:heatmap}a represents \textit{text} modality and right side represents the \textit{visual} modality. The heatmaps represent the contributing features for the classification of utterances. Each cell (\textit{i,j}) of Figure \ref{fig:heatmap} signifies the weights of utterance `\textit{j}' for the classification of utterance `\textit{i}' of the pair-wise modality matrices. For example, for the classification of utterance `\textit{u4}' in Figure \ref{fig:heatmap}a, model puts more focus on the textual features of `\textit{u2}' and `\textit{u6}' than others and more-or-less equal focus on the visual features of all the utterances.

\subsection{Comparative Analysis}
We compare our proposed approach against various existing systems \cite{Nojavanasghari:2016:DMF:2993148.2993176,mvlstm:2016,zadeh2017tensor,zadeh2018multi-sdk,mfn:zadeh:aaa1:2017,zadeh2018acl,W18-3301} that made use of the same datasets. A comparative study is shown in Table \ref{tab:results-compare}. We report the results of the top three existing systems (as reported in \citealt{zadeh2018acl}) for each case. In emotion classification, the proposed multi-task learning framework reports the best F1-score of 78.6\% as compared to the 76.3\% and Weighted Accuracy of 62.8\% as compared to the 62.3\% of the state-of-the-art. Similarly, for sentiment classification, the state-of-the-art system reports 77.0\% F1-score and 76.9\% accuracy value in the multi-task framework. In comparison, we obtain the best F1-score and accuracy value of 78.8\% and 80.4\%, respectively, i.e., an improvement of 1.8\% and 3.5\% over the state-of-the-art systems.

During analysis, we make an important observation. Small improvements in performance do not reveal the exact improvement in the number of instances. Since there are more than 4.6K test samples, even the 
improvement by one point reflects that the system improves its predictions for 46 samples.

We also perform test-of-significance (\textit{T}-test) and observe that the obtained results are statistically significant \textit{w.r.t.} the state-of-the-art and proposed single-task results with \textit{p}-values$<0.05$. 

\input{error_analysis}

\subsection{STL v/s MTL framework}
In this section, we present our analysis \textit{w.r.t.} single-task and multi-task frameworks. Table \ref{tab-error-extensive} lists a few example cases where the proposed multi-task learning framework shows how it yields better performance for both sentiment and emotion, while the single-task framework finds it non-trivial for the classification. For example, first utterance has gold sentiment label as negative which was misclassified by STL framework. However, the MTL framework improves this by correctly predicting `\textit{positive}'. Similarly, in emotion analysis STL predicts three emotions i.e. \textit{disgust}, \textit{happy} and \textit{sad}, out of which only one emotion (\textit{disgust}) matches the gold emotions of \textit{anger} and \textit{disgust}. In comparison, MTL predicts four emotions (i.e. \textit{anger}, \textit{disgust}, \textit{happy} and \textit{sad}) for the same utterance. The precision (2/4) and recall (2/2) for MTL framework is better than the precision (1/3) and recall (1/2) for the STL framework. These analyses suggest that the MTL framework, indeed, captures better evidences than the STL framework.  

In the second example, knowledge of sentiment helps in identifying the correct emotion label in the MTL framework. For the gold sentiment (\textit{positive}) and emotion (\textit{happy} and \textit{sad}) labels, STL correctly classifies one emotion (i.e. \textit{sad}), but fails to predict the other emotion (i.e. \textit{happy}). In addition, it misclassifies another emotion (i.e. \textit{anger}). Since, gold label \textit{happy} corresponds to the \textit{positive} scenario and predicted label \textit{anger} is related to negative, knowledge of sentiment is a crucial piece of information. Our MTL framework identifies this relation and leverage the predicted sentiment for the classification of emotion i.e. \textit{positive} sentiment assists in predicting \textit{happy} emotion. This is an example of inter-dependence between the two related tasks and the MTL framework successfully exploits it for the performance improvement. 

We also observe that the system puts comparatively more focus on some classes in MTL framework than the STL framework. As an instance, MTL predicts `\textit{anger}' class for 1173 utterances, whereas STL predicts it for 951 utterances (1063 anger utterances in the gold dataset). Further, we observe contrasting behavior for the `\textit{sad}' class, where MTL predicts 1618 utterances as `\textit{sad}' compared to the 2126 utterances of STL. For `\textit{disgust}' and `\textit{happy}' classes, both STL and MTL frameworks predict the approximately equal number of utterances. 

Further, we observe that MTL performs poorly for the `\textit{fear}' and `\textit{surprise}' classes, where it could not predict a significant number of utterances. A possible reason would be the under-representation of these instances in the given dataset. 

\section{Conclusion}
\label{sec:con}
In this paper, we have proposed a deep multi-task framework that aims to leverage the inter-dependence of two related tasks, i.e., multi-modal sentiment and emotion analysis. Our proposed approach learns a joint-representation for both the tasks as an application of GRU based inter-modal attention framework. We have evaluated our proposed approach on the recently released benchmark dataset on multi-modal sentiment and emotion analysis (MOSEI). Experimental results suggest that sentiment and emotion assist each other 
when learned in a multitask framework. We have compared our proposed approach against the various existing systems and observed that multi-task framework attains higher performance for all the cases.          

In the future, we would like to explore the other dimensions to our multi-task framework, e.g., Sentiment classification \& intensity prediction, Emotion classification \& intensity prediction and all the four tasks together. \\

\noindent {\large \bf Acknowledgement}

\vspace{0.5em}
\noindent Asif Ekbal acknowledges the Young Faculty Research Fellowship (YFRF), supported by Visvesvaraya Ph.D. scheme for Electronics and IT, Ministry of Electronics and Information Technology (MeitY), Government of India, being implemented by Digital India Corporation (formerly Media Lab Asia). 

%% file: dataset.tex
\begin{table}
\centering
\resizebox{0.32\textwidth}{!}
{
\begin{tabular}{|l|ccc|}
\hline
\textbf{Statistics} & Train & Dev & Test \\ \hline \hline
\textit{\#Videos}  &2250 & 300 & 679 \\
\textit{\#Utterance}  & 16216 & 1835 & 4625 \\
\textit{\#Positive}  & 11499 & 1333 & 3281\\
\textit{\#Negative}  & 4717 & 502 & 1344 \\
\textit{\#Anger}  & 3506 & 334 & 1063 \\
\textit{\#Disgust}  & 2946 & 280 & 802 \\
\textit{\#Fear}  & 1306 & 163 & 381 \\
\textit{\#Happy}  & 8673 & 978 & 2484 \\
\textit{\#Sad}  & 4233 & 511 & 1112 \\
\textit{\#Surprise}  & 1631 & 194 & 437 \\
\textit{\#Speakers}  & \multicolumn{3}{c|}{1000}\\ \hline
\end{tabular}
}
\caption{Dataset statistics for CMU-MOSEI. Each utterance contains multi-modal information.} 
\label{tab:dataset}
\end{table}

%% file: results-multi-single.tex
\begin{table*}[!ht]
\centering
\resizebox{1\textwidth}{!}
{
\begin{tabular}{c|c||ccc|ccc|c||ccc|ccc|c}
\bf \multirow{2}{*}{Tasks} & & \multicolumn{7}{c||}{\bf F1-score} & \multicolumn{7}{c}{\bf Acc (Sentiment) \& Weighted-Acc (Emotion)} \\ \cline{3-16}
& & \bf T & \bf A & \bf V & \bf T+V & \bf T+A & \bf A+V & \bf T+A+V & \bf T & \bf A & \bf V & \bf T+V & \bf T+A & \bf A+V & \bf T+A+V \\ \hline \hline
\multirow{2}{*}{Sent} & STL & 75.1 & 67.9 & 66.3 & 77.0 & 76.5 & 69.6 & 77.6 & 78.2 & 74.8 & 75.8 & 79.4 & 79.7 & 76.6 & 79.8 \\ 
& MTL & 77.5 & 72.1 & 69.1 & 78.7 & 78.6 & 75.8 & 78.8 & 79.7 & 75.7 & 76.5 & 80.4 & 80.2 & 77.4 & 80.5 \\ \hline
\multirow{2}{*}{Emo} & STL & 75.9 & 72.3 & 73.6 & 77.5 & 76.8 & 76.0 & 77.7 & 58.0 & 56.7 & 53.7 & 60.1 & 59.6 & 58.0 & 60.8 \\ 
 & MTL & 76.9 & 74.6 & 75.4 & 78.5 & 77.6 & 77.0 & 78.6 & 60.2 & 56.2 & 57.5 & 62.5 & 60.5 & 59.3 & 62.8 \\
\end{tabular}
}
\caption{Single-task learning (STL) and Multi-task (MTL) learning frameworks for the proposed approach. T: Text, V: Visual, A: \textit{Acoustic}. Weighted accuracy as a metric is chosen due to unbalanced samples across various emotions and it is also in line with the other existing works \cite{zadeh2018acl}.}
\label{tab:mtl:stl}
\end{table*}

%% file: results.tex
\begin{figure*}[ht!]
\centering
\resizebox{1.0\textwidth}{!}
{
\begin{tikzpicture}

\begin{axis}[
    ybar,
    name = plot1,
    width=11cm,
 	height =3.6cm,
    ylabel={F1-Score},
    ymin=65, ymax=80,
    symbolic x coords={T,A,V,T+V,T+A,A+V,T+A+V},
    xtick=data,
    ymajorgrids=true,
    grid style=dashed,
]

\addplot[color=blue, fill=blue!50,] coordinates {(T,75.1) (A,67.9) (V,66.3) (T+V,77.0) (T+A,76.5) (A+V,69.6) (T+A+V,77.67) }; 
\addplot[color=red, fill=red!50,postaction={pattern=north east lines}] coordinates {(T,77.5) (A,72.1) (V,69.1) (T+V,78.7) (T+A,78.6) (A+V,75.8) (T+A+V,78.86) }; 
\end{axis}

\begin{axis}[
    ybar,
    name = plot2,
    xshift=0.2cm,
    at=(plot1.right of south east), anchor=left of south west,
 	width=11cm,
 	height =3.6cm,
    ylabel={F1-Score},
    ylabel near ticks,
    ymin=70, ymax=80,
    symbolic x coords={T,A,V,T+V,T+A,A+V,T+A+V},
    xtick=data,
    ymajorgrids=true,
    yticklabel pos=right,
    grid style=dashed,
]

\addplot[color=blue, fill=blue!50,] coordinates {(T,75.9) (A,72.3) (V,73.6) (T+V,77.5) (T+A,76.8) (A+V,76.0) (T+A+V,77.7) }; 
\addplot[color=red, fill=red!50,postaction={pattern=north east lines}] coordinates {(T,76.9) (A,74.6) (V,75.4) (T+V,78.5) (T+A,77.6) (A+V,77.0) (T+A+V,78.6) }; 
\end{axis}

\begin{axis}[
    ybar,
    name = plot3,
    at=(plot1.below south west), anchor=above north west,
 	height =3.6cm,
 	width=11cm,
    ylabel={Accuracy},
    xlabel={Sentiment},
    ymin=74, ymax=81,
    symbolic x coords={T,A,V,T+V,T+A,A+V,T+A+V},
    xtick=data,
    ymajorgrids=true,
    grid style=dashed,
]

\addplot[color=blue, fill=blue!50,] coordinates {(T,78.2) (A,74.8) (V,75.8) (T+V,79.4) (T+A,79.7) (A+V,76.6) (T+A+V,79.8) }; 
\addplot[color=red, fill=red!50,postaction={pattern=north east lines}] coordinates {(T,79.7) (A,75.7) (V,76.5) (T+V,80.4) (T+A,80.2) (A+V,77.4) (T+A+V,80.5) }; 
\end{axis}

\begin{axis}[
    ybar,
    area legend,
    name = plot4,
    xshift=0.2cm,
    at=(plot3.right of south east), anchor=left of south west,
 	width=11cm,
 	height =3.6cm,
    ylabel={Weighted Accuracy},
    ylabel near ticks,
    xlabel={Emotion},
    ymin=52, ymax=65,
    symbolic x coords={T,A,V,T+V,T+A,A+V,T+A+V},
    xtick=data,
    ymajorgrids=true,
    yticklabel pos=right,
    grid style=dashed,
    legend style={font=\scriptsize, legend pos= north west, legend columns=2,  legend style={row sep=0.5pt}},
]

\addplot[color=blue, fill=blue!50,] coordinates {(T,58.0) (A,56.7) (V,53.7) (T+V,61.0) (T+A,59.6) (A+V,58.0) (T+A+V,60.8) }; \addlegendentry{STL}
\addplot[ color=red, fill=red!50, postaction={pattern=north east lines}] coordinates {(T,60.2) (A,56.2) (V,57.5) (T+V,62.5) (T+A,60.5) (A+V,59.3) (T+A+V,62.8) }; \addlegendentry{MTL}

\end{axis}
\end{tikzpicture}
}
\caption{Single-task learning (STL) and Multi-task (MTL) learning frameworks for the proposed approach. 
}
\label{mtl-stl-graph}
\end{figure*}
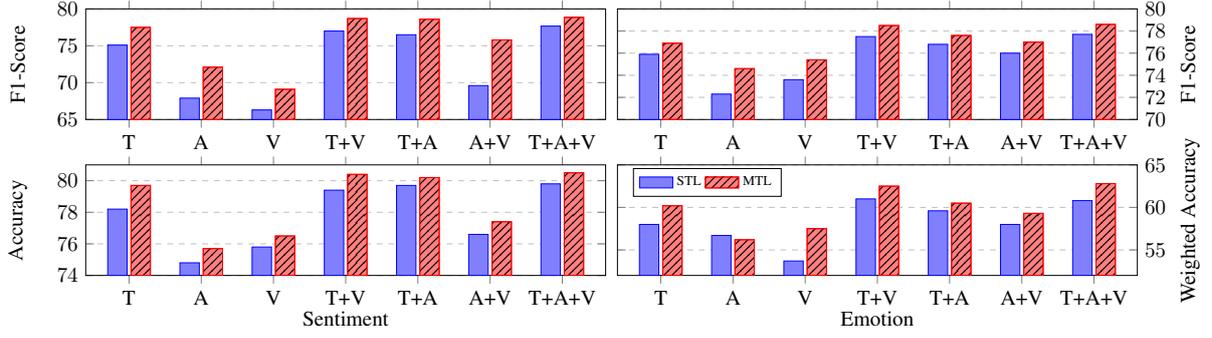



%% file: table-attention.tex
\begin{table*}
\centering
\resizebox{1.0\textwidth}{!}
{
\begin{tabular}{l|p{30em}||c|c||l|p{7em}}
& & \multicolumn{2}{c||}{\bf Sentiment} & \multicolumn{2}{c}{\bf Emotion} \\ \cline{3-6}
& \bf Utterances & \bf Actual & \bf MTL & \multicolumn{1}{c|}{\bf Actual} & \multicolumn{1}{c}{\bf MTL} \\ \hline \hline
1 & \textit{line hello my name is sarah and i will be doing my video opinion on the movie shall we dance uhh starring jennifer} & Pos & Pos & Anger &  Anger	\\  \hline

2 & \textit{richard gere and susan umm you i really didn't enjoy this movie at all it kinda boring for}	& Neg	& Neg	&	Anger, Disgust	&	Anger, Disgust, \textcolor{red}{ Happy, Sad}	\\ \hline

3 & \textit{for umm it kinda felt as if there were parts in there they}	&	Pos	&	\textcolor{red}{Neg}	&	No class & \textcolor{red}{Anger, Disgust, Happy, Sad}	\\ \hline

4 & \textit{they just put in there to kinda pass the time on basically the movie is about umm richard character and him being a}	 & Pos & Pos & Happy	&	\textcolor{red}{Anger, Disgust}, Happy, \textcolor{red}{Sad}	\\ \hline

5 & \textit{umm looking for some some stutter extra sizzle to add into his life he meets up with a dance instructor who is played by jennifer lopez and basically she convinces him to sign up for some ballroom he gets into it he enjoys it a lot but still a secret from his family} & Pos & Pos	&	Anger	&	Anger \textcolor{red}{Disgust, Happy, Sad}	\\ \hline

6 & \textit{family he is trying to cope with having this this stutter}	&	Neg	&	Neg	&	Anger, Disgust	&	Anger, Disgust, \textcolor{red}{Happy, Sad}	\\ \hline

 \hline
\end{tabular}
}
\caption{Example video for heatmap analysis of the contextual inter-modal (CIM) attention mechanism of the proposed MTMM-ES framework. Figure \ref{fig:heatmap} depicts the heatmaps for the above video.}
\label{tab-heatmap-video-example}
\end{table*}

%% file: all_attention.tex
\begin{figure*}[!ht]       
	\begin{center}
    \subfloat[\textit{Softmax attention weights $N_1$ \& $N_2$ for TV}.\label{video-text}]
       { \includegraphics[width=0.3\textwidth, height=3.0cm]{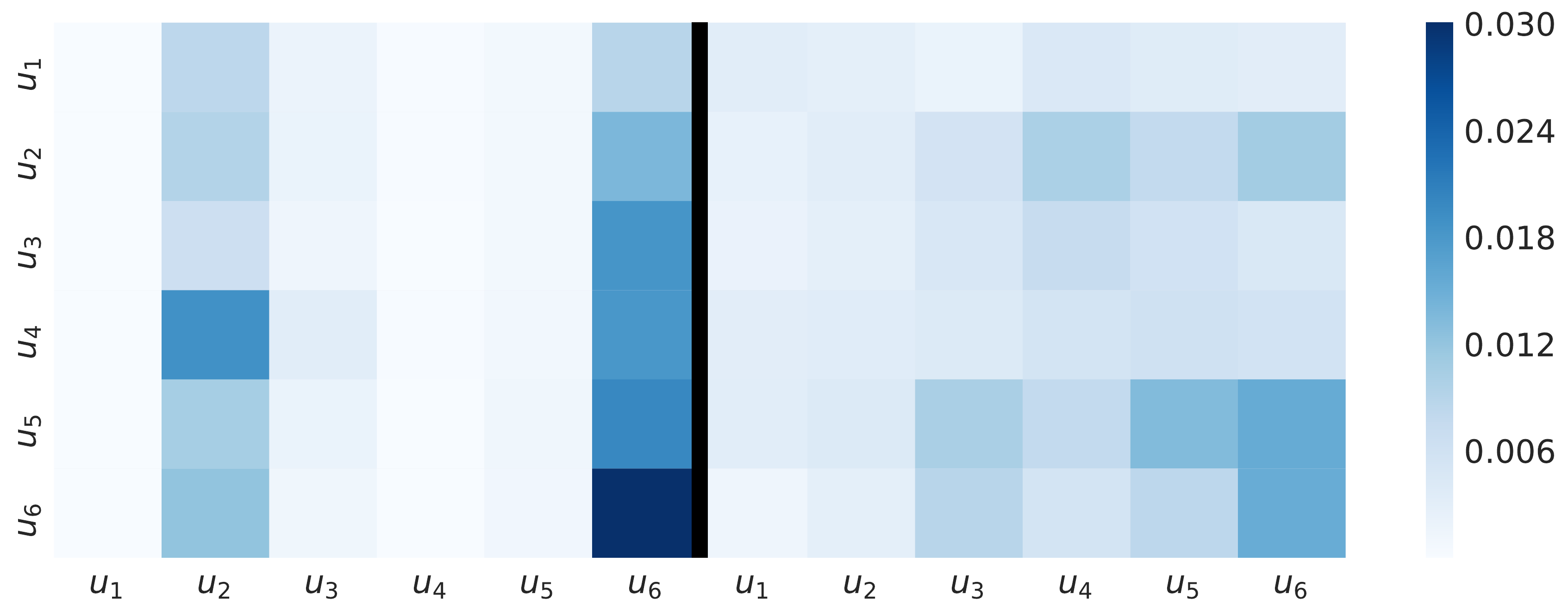}
       }
       \hspace{4pt}
       \subfloat[\textit{Softmax attention weights $N_1$ \& $N_2$ for AV}.\label{audio-video}]
       {
        \includegraphics[width=0.3\textwidth, height=3.0cm]{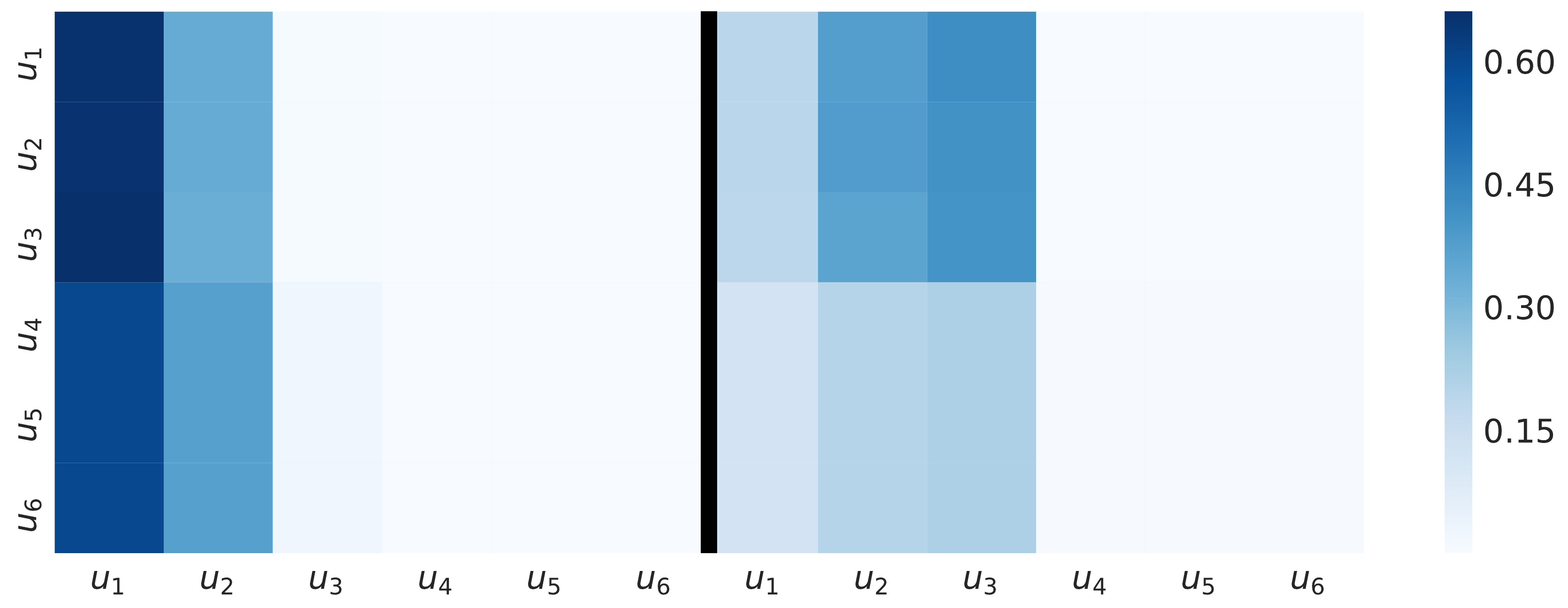}
       }
     \hspace{4pt}
       \subfloat[\textit{Softmax attention weights $N_1$ \& $N_2$ for TA}.\label{text-audio}]
       {
        \includegraphics[width=0.3\textwidth, height=3.0cm]{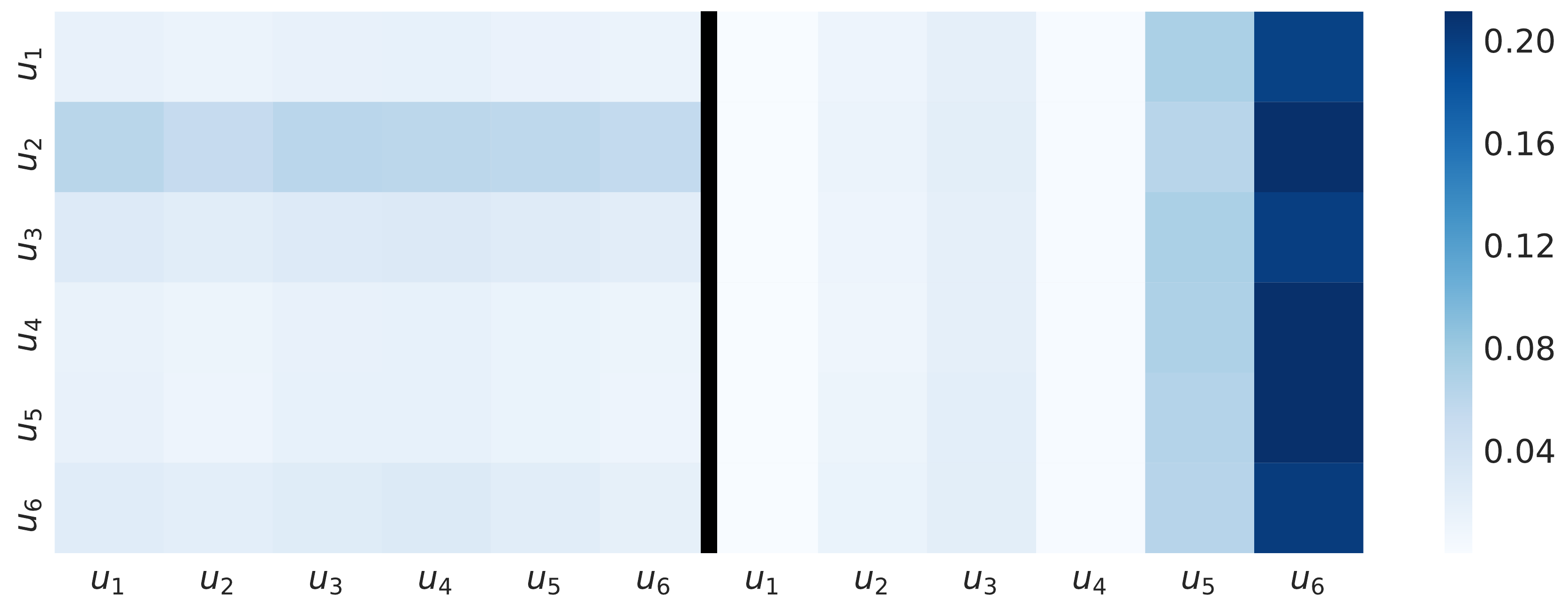}
       }

	\end{center}  
    \caption{(a), (b) \& (c): Pair-wise softmax attention weights $N_1$ \& $N_2$ of \textit{visual-text}, \textit{acoustic-visual} \& \textit{text-acoustic} for multi-task learning framework. Solid line at the center represents boundary of $N_1$ \& $N_2$. The heatmaps represent attention weights of a particular utterance with respect to other utterances in $N_1$ \& $N_2$. Each cell ($i,j$) of the heatmap signifies the weights of utterance `$j$' for the classification of utterance `$i$' of the pair-wise modality matrices, hence, assists in predicting the labels concisely by incorporating contextual inter-modal information. 
}
\label{fig:heatmap}
 
	\end{figure*}

%% file: results-compare-new.tex
\begin{table*}[ht]
\centering
\resizebox{0.98\textwidth}{!}
{
\begin{tabular}{l||cc|cc|cc|cc|cc|cc|cc||cc}
 & \multicolumn{14}{c||}{\bf \em Emotion}& \multicolumn{2}{c}{\multirow{1}{*}{\bf \em Sentiment}} \\ \cline{2-17}

& \multicolumn{2}{c|}{\bf \em Anger} & \multicolumn{2}{c|}{\bf \em Disgust}& \multicolumn{2}{c|}{\bf \em Fear} & \multicolumn{2}{c|}{\bf \em Happy}& \multicolumn{2}{c|}{\bf \em Sad} & \multicolumn{2}{c|}{\bf \em Surprise}& \multicolumn{2}{c||}{\bf \em Average$^\dagger$} & \multicolumn{2}{c}{\bf \em } \\
 
System & \textit{F1} & \textit{W-Acc}  & \textit{F1} & \textit{W-Acc}  & \textit{F1} & \textit{W-Acc} & \textit{F1} & \textit{W-Acc} &  \textit{F1} & \textit{W-Acc} & \textit{F1} & \textit{W-Acc}& \textit{F1} & \textit{W-Acc} & \textit{F1} & \textit{Acc}  \\\hline \hline
\newcite{W18-3301} & - & - & - & - & - & - & - & - & - & - & - & - & - & - & \textit{63.2} & \textit{60.0} \\
\newcite{mfn:zadeh:aaa1:2017}$^\star$ & - & - & 71.4 & 65.2 & \bf 89.9 & - & - & - & 60.8 & - & 85.4 & 53.3 & - & - & 76.0 & 76.0 \\
\newcite{Nojavanasghari:2016:DMF:2993148.2993176}$^\star$ & 71.4 & - & - & 67.0 & - & - & - & - & - & - & - & - & - & - & - & - \\
\newcite{mvlstm:2016}$^\star$ & - & 56.0 & - & - & - & - & - & - & - & - & - & - & - & - & 76.4 & 76.4 \\
EF-LSTM \cite{zadeh2018acl}$^\star$ & - & - & - & - & - & 56.7 & - & 57.8 & - & 59.2 & - & - & - & - & - & - \\
TFN \cite{zadeh2017tensor}$^\star$ & - & 60.5 & - & - & - & - & 66.6 & 66.5 & - & 58.9 & - & 52.2 & - & - & - & - \\
Random Forest \cite{random:forest:breiman2001}$^\star$ & 72.0 & - & 73.2 & - & \bf 89.9 & - & - & - & 61.8 & - & 85.4 & - & - & - & - & - \\
SVM \cite{zadeh:mosi}$^\star$ & - & - & - & - & - & 60.0 & - & - & - & - & - & - & - & - & - & - \\
\newcite{zadeh2018multi-sdk}$^\star$ & - & - & - & - & - & - & \bf 71.0 & - & - & - & - & - & - & - & - & - \\
\newcite{zadeh2018acl} & 72.8 & 62.6 & 76.6 & 69.1 & \bf 89.9 &  62.0 & 66.3 & 66.3 & 66.9 & 60.4 & 85.5 &  53.7 & 76.3 &  62.3 & 77.0 & 76.9 \\
\hline
Proposed (Single-task learning) &  75.6 &  64.5 &  81.0 &  72.2 & 87.7 & 51.5 & 59.3 & \bf 61.6 &  67.3 & \bf 65.4 & \bf 86.5 & 53.0  & 76.2 & 61.3 & 77.6 & 79.8 \\ 
\bf Proposed (Multi-task learning) & \bf 75.9 & \bf 66.8 & \bf 81.9 & \bf 72.7 &  87.9 & \bf 62.2 & 67.0 & 53.6 & \bf 72.4 & 61.4 & 86.0 & \bf 60.6 & \bf 78.6 & \bf 62.8 & \bf 78.8 & \bf 80.5 \\ \hline
Significance \textit{T}-test \textit{w.r.t.} SOTA & \bf \textit{-} & \bf \textit{-} & \bf \textit{-} & \bf \textit{-} &  \textit{-} & \textit{-} & \bf \textit{-} & \bf \textit{-} & \bf \textit{-} & \bf \textit{-} & \em \textit{-} & \bf \textit{-} & \bf \textit{0.0240} & \bf \textit{0.0420} & \bf \textit{0.0012} & \bf \textit{0.0046} \\
Significance \textit{T}-test \textit{w.r.t.} STL & \bf \textit{-} & \bf \textit{-} & \bf \textit{-} & \bf \textit{-} &  \textit{-} & \textit{-} & \bf \textit{-} & \bf \textit{-} & \bf \textit{-} & \bf \textit{-} & \em \textit{-} & \bf \textit{-} & \bf \textit{ 0.0171} & \bf \textit{0.0312} & \bf \textit{0.0015} & \bf \textit{0.0278} \\ 
\end{tabular}
}
\caption{Comparative results: Proposed multi-task framework attains better performance as compared to the state-of-the-art (SOTA) systems in both the tasks i.e. emotion recognition (average) and sentiment analysis. $^\star$Values are taken from \newcite{zadeh2018acl}. $^\dagger$Six-class average. Significance \textit{T}-test ($<0.05$). STL: Single-task learning. 
}
\label{tab:results-compare}
\end{table*}

%% file: error_analysis.tex
\begin{table*}
\centering
\resizebox{0.96\textwidth}{!}
{
\begin{tabular}{l|p{25em}||c|c|c||l|l|l}
& & \multicolumn{3}{c||}{\bf Sentiment} & \multicolumn{3}{c}{\bf Emotion} \\ \cline{3-8}
& \bf Utterances & \bf Actual & \bf STL & \bf MTL & \bf Actual & \bf STL & \bf MTL \\ \hline \hline
1 & \textit{richard gere and susan umm you i really didn't enjoy this movie at all it kinda boring for} & Neg & \textcolor{red}{Pos} & Neg & Anger, Disgust & Disgust, \textcolor{red}{Happy, Sad}	& Anger, Disgust, \textcolor{red}{Happy, Sad}	\\  \hline
2 & \textit{we look forward to cooperating with the new government as it works to make progress on a wide range of issues including further democratic reforms promotion of human rights economic development and national reconciliation}	& Pos	& Pos	& Pos	&	Happy, Sad	&	\textcolor{red}{Anger}, Sad	&	Happy, Sad	\\ \hline
3 & \textit{laughter and applause still there though..}	&	Pos	&	\textcolor{red}{Neg} & Pos	&	Happy	&	Happy, \textcolor{red}{Surprise} &	Happy	\\ \hline
4 & \textit{is in love with some other person so you know the story}	& Neg &	\textcolor{red}{Pos} & Neg & Anger, Disgust, Sad	& Disgust, \textcolor{red}{Happy}, Sad &	Anger, Disgust, Sad	\\ \hline
5 & \textit{i can say unfortunately i don't think it's a serious program} & Neg & \textcolor{red}{Pos} &	Neg	&	Disgust, Sad, Surprise	&	\textcolor{red}{Anger, Happy}, Sad	&	\textcolor{red}{Anger}, Disgust, \textcolor{red}{Happy}, Sad	\\ \hline
6 & \textit{the last administration bought into just as much as this one does unfortunately}	&	Neg	&	\textcolor{red}{Pos}	&	Neg	&	Anger, Disgust, Sad	&	Anger, \textcolor{red}{Happy}	&	Anger, Disgust, \textcolor{red}{Happy}, Sad	\\ \hline
7 & \textit{it's just too great of a risk and it is socially unacceptable}	& Neg	&	\textcolor{red}{Pos} &	Neg	&	Anger, Disgust, Happy	&	Anger, Happy	&	Anger, Disgust, Happy	\\ \hline
8 & \textit{had a robot here at hopkins since the year longer than most institutions in this country and around the world we} &	Pos	&	Pos	&	Pos	&	Happy	&	Happy, \textcolor{red}{Sad, No class} & Happy	\\ \hline
9 & \textit{in total we spent hundreds of hours on the ground on site watching these leaders in action}	&	Pos	&	Pos	&	Pos	&	No class	&	\textcolor{red}{Happy} & No class	\\
 \hline
\end{tabular}
}
\caption{Comparison with multi-task learning and single-task learning frameworks. Few error cases where multi-task learning framework performs better than the single-task framework. \textit{First utterance}: Improved MTL (\textit{Pre: 0.5, Rec: 1.0}) performance over STL (\textit{Pre: 0.3, Rec: 0.5}). \textit{Second utterance:} Sentiment (i.e. \textit{Pos}) assists in emotion classification (i.e. \textit{Happy}). Red color represents error in classification. }
\label{tab-error-extensive}
\end{table*}